\documentclass[10pt,twocolumn,letterpaper]{article}

\usepackage[pagenumbers]{cvpr} 

\definecolor{cvprblue}{rgb}{0.21,0.49,0.74}
\usepackage[pagebackref,breaklinks,colorlinks,allcolors=cvprblue]{hyperref}
\usepackage[accsupp]{axessibility}
\def\paperID{*****} 
\def\confName{CVPR}
\def\confYear{2026}

\title{FAGER: Factually Grounded Evaluation and Refinement of Text-to-Image Models}

\author{
Youngsun Lim\\
Boston University\\
{\tt\small youngsun@bu.edu}
\and
Cusuh Ham\\
Adobe\\
{\tt\small ham@adobe.com}
\and
Pin-Yu Chen\\
IBM Research\\
{\tt\small pin-yu.chen@ibm.com}
\and
Deepti Ghadiyaram\\
Boston University\\
{\tt\small dghadiya@bu.edu}
}

\begin{document}
\maketitle
\begin{abstract}
Existing text-to-image (T2I) evaluation metrics mainly assess whether generated images align with information explicitly stated in the prompt, but often fail to capture factual requirements that are implicit, externally grounded, or identity-defining. As a result, they are not well suited for evaluating factual correctness in prompts involving scientific knowledge, historical facts, products, or culture-specific concepts. We propose \textbf{FA}ctually \textbf{G}rounded \textbf{E}valuation and \textbf{R}efinement (\textbf{FAGER}), an agentic framework that evaluates whether generated images correctly reflect visually verifiable facts grounded in or implied by the prompt, while also providing actionable feedback for improvement. FAGER first constructs a structured factual rubric by combining LLM-based fact proposal with reference-guided visual fact extraction and verification, then converts the rubric into question--answer pairs for VLM-based evaluation. To validate FAGER as a factuality metric, we introduce a \textbf{Factual A/B test}, which measures whether a metric prefers factual reference images over corresponding generated images. Across five datasets spanning science, history, products, culture, and knowledge-intensive concepts, FAGER consistently outperforms prior metrics on this test. We further show that FAGER can be used to refine T2I outputs in a fully training-free manner, yielding substantial factuality gains across datasets. 
Code and data are available at \url{https://github.com/SGT-LIM/FAGER}.
\end{abstract}    
\section{Introduction}
\label{sec:intro}

\begin{figure}[t]
    \centering
    \includegraphics[width=1.0\linewidth]{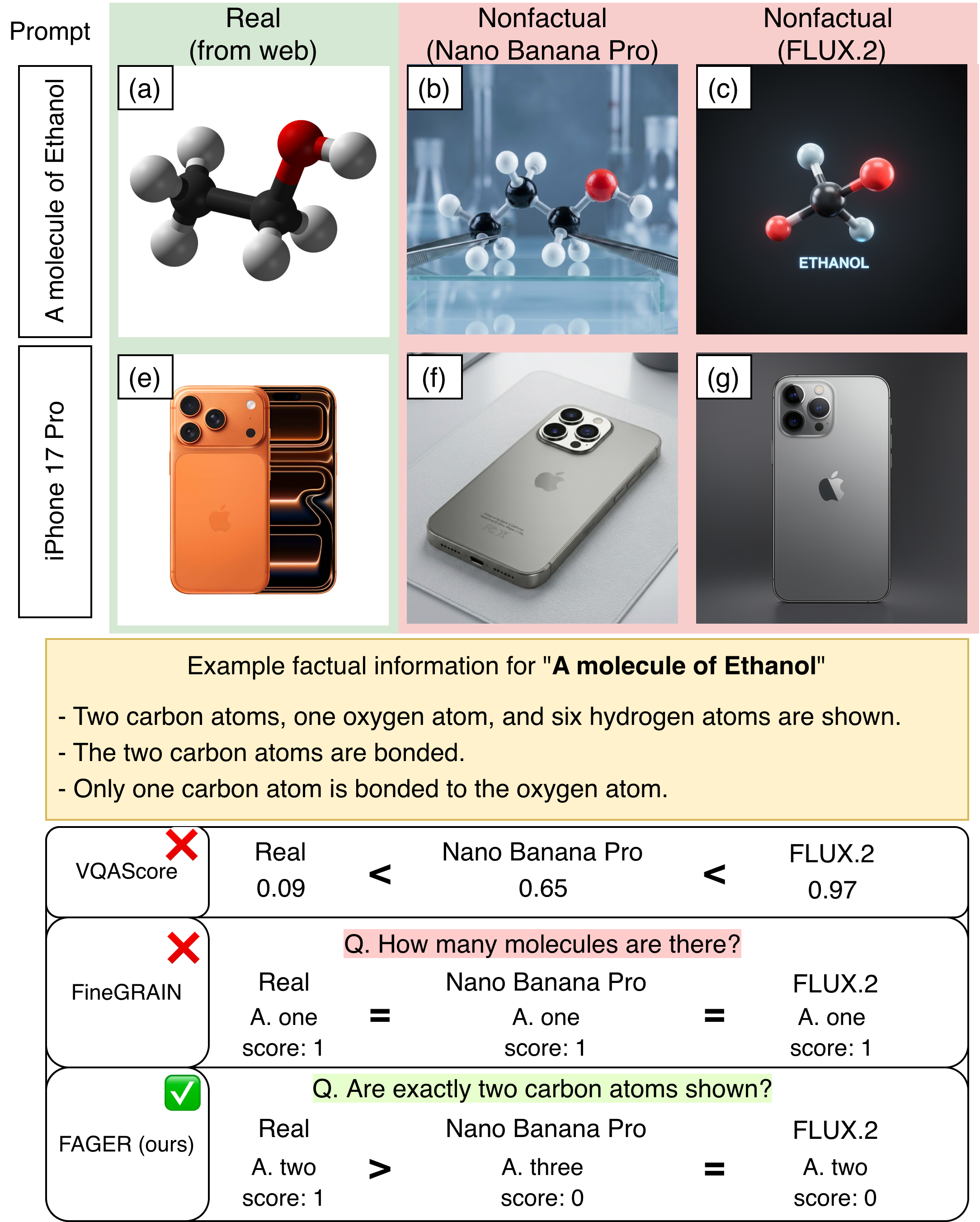}
    \caption{\textbf{Overview of real and nonfactual images and their evaluations by different T2I metrics.} Existing metrics such as VQAScore~\cite{lin2024evaluating} and FineGRAIN~\cite{hayes2025finegrain} fail to properly assess facts implied by the prompt. In the ethanol example, VQAScore assigns higher scores to nonfactual images, while FineGRAIN focuses on a prompt-explicit question that does not distinguish factual correctness. In contrast, FAGER evaluates prompt-grounded facts through targeted question answering, correctly identifying the factual reference image and penalizing nonfactual generations.}
    \label{fig:1}
\end{figure}

Recent advances in text-to-image (T2I) models have significantly improved their ability to generate visually appealing and stylistically rich images~\cite{blackforestlabs2024flux1dev, flux-2-2025, sd35_huggingface_2024, baldridge2024imagen}. 
Yet, as illustrated in \cref{fig:1}, current models still struggle to generate factual images across diverse domains, including scientific knowledge (\eg, ``A molecule of Ethanol'') and emerging or novel concepts (\eg, ``iPhone 17 Pro'').
Importantly, these failures cannot be easily resolved by simply providing more detailed prompts~\cite{kamath2025geneval}. 
Furthermore, such an approach would require users to explicitly specify all relevant factual details, which is rarely realistic in practice. For example, users usually write ``a molecule of ethanol,'' but typically do not specify that it should contain two carbon atoms, one oxygen atom, and six hydrogen atoms arranged in the correct structure.

On the other hand, as described in \cref{fig:1}, most existing metrics rely heavily on the input prompt and are based on visual question answering (VQA) to visually verify the entities, object counts, and spatial relations~\cite{hayes2025finegrain, lin2024evaluating, hu2023tifa}. We argue that there is a need for evaluation metrics that capture the explicitly described information in the prompts and the implicitly accurate facts.

To this end, we propose \textbf{FA}ctually \textbf{G}rounded \textbf{E}valuation and \textbf{R}efinement (FAGER),
an agentic framework that assesses whether generated images correctly reflect visually verifiable facts grounded in or implied by the prompt, while also providing actionable feedback signals for improvement (\cref{fig:2}). 
Our approach leverages the prior knowledge of a large language model (LLM) to extract facts from prompts. 
Next, we refine the LLM-proposed facts using reference images sourced from existing benchmarks and Google Image Search, and convert the refined facts into high-quality question-answer (QA) sets. These QA sets are then used by a vision-language model (VLM) to evaluate generated images. 
Through experiments, we demonstrate that our method better captures factual correctness compared to existing metrics. 
Furthermore, we show that our evaluation outputs can be used as feedback to iteratively improve image generation, leading to more factually accurate results. 
Our approach is training-free, generalizes across domains, and remains flexible: users can dynamically generate QA sets for arbitrary prompts and use our pipeline to both evaluate and improve generated images.
\section{Related works: Existing T2I metrics}
\label{sec:related}

Early T2I evaluation metrics primarily measured the alignment between generated images and input prompts. For example, CLIPScore~\cite{hessel2021clipscore} evaluates image--text similarity using CLIP embeddings, while DALL-Eval~\cite{cho2023dall} assesses prompt following through a fixed set of evaluation criteria. Although these methods are useful for measuring coarse alignment, they are limited by the inherent weaknesses of CLIP-based similarity, such as poor counting ability and weak compositional reasoning.


With the rise of stronger vision-language models, a new class of T2I evaluation methods has emerged that formulates evaluation as visual question answering (VQA). TIFA~\cite{hu2023tifa} decomposes prompts into semantic units, generates question--answer pairs, and evaluates whether a VQA model can correctly answer them from the generated image. VQ$^2$~\cite{yarom2023you} similarly extracts key information from prompts and evaluates image support through generated questions. More recently, FineGRAIN~\cite{hayes2025finegrain} broadens the evaluation scope by introducing 27 fine-grained failure categories for structured T2I assessment.
Like these methods, FAGER also uses question answering for evaluation, but goes beyond prompt-explicit content by using reference-guided rubric construction to target visually verifiable facts implied by the prompt, and further supports downstream refinement through edit and regeneration feedback.

Existing metrics still primarily measure whether the generated image matches information explicitly stated in the prompt. As a result, they often fail to assess factual requirements that are implicit or identity-defining. T2I-FactualBench~\cite{huang2024t2i} takes an important step beyond prompt-only evaluation by leveraging reference images to evaluate aspects not explicitly described in the prompt. However, it is still not designed to distinguish ``facts'' from commonsense expectations or conventional depictions. As we discuss in Sec.~\ref{sec:3.1}, such properties should not necessarily be treated as facts, which limits the suitability of prior metrics for rigorous factuality evaluation.

\section{Methodology}

\begin{figure*}[t]
    \centering
    \includegraphics[width=\textwidth]{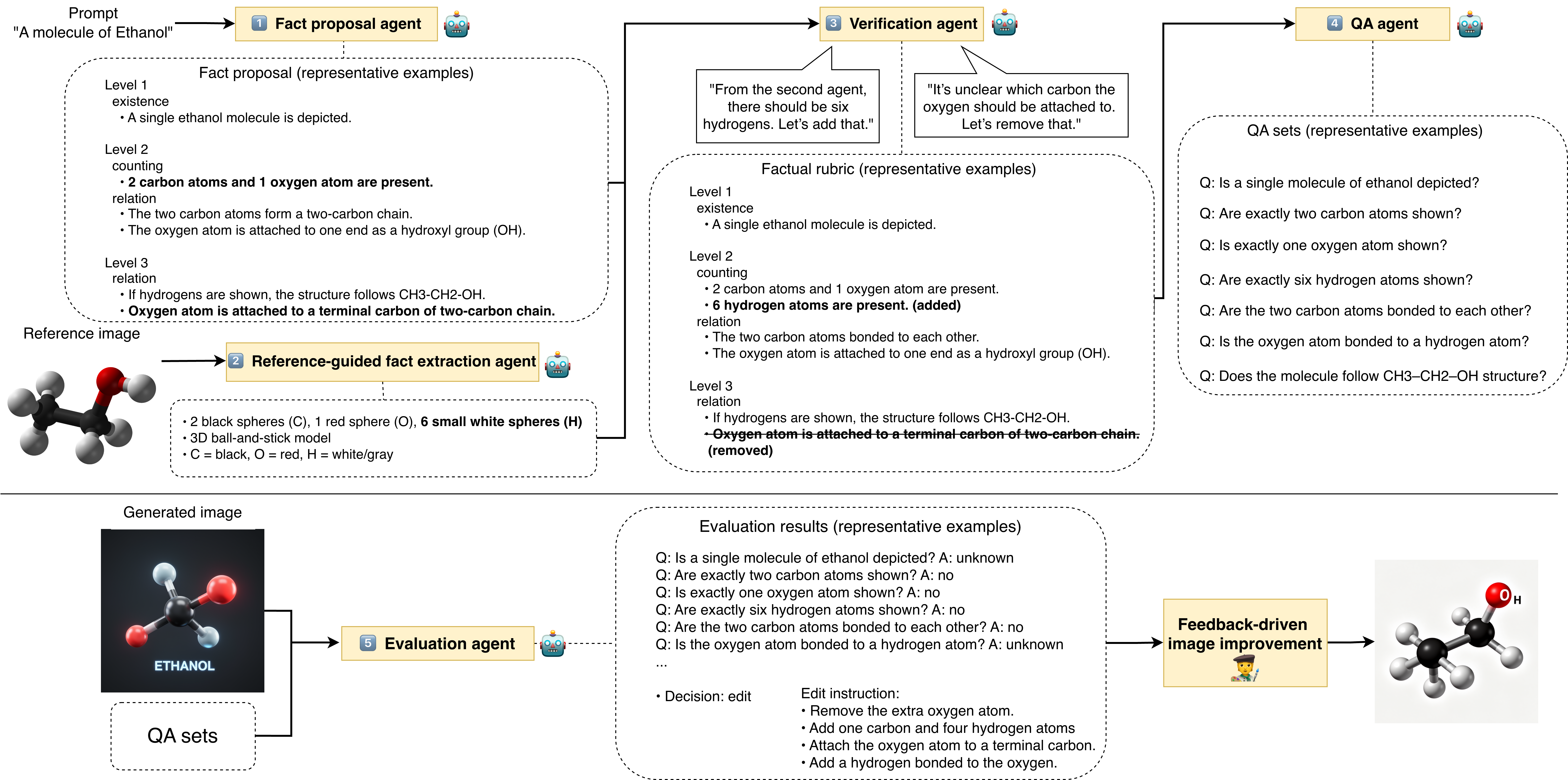}
    \caption{\textbf{Overview of the proposed FAGER pipeline on the prompt ``A molecule of Ethanol.''}
    The \textbf{fact proposal agent} generates structured candidate facts from the prompt using prior knowledge.
    In parallel, the \textbf{reference-guided fact extraction agent} extracts visually observable facts from a reference image.
    The \textbf{verification agent} then consolidates the two sources into a verified \textbf{factual rubric}, adding missing factual requirements and removing unsupported ones.
    The \textbf{QA agent} converts the verified rubric into question--answer pairs, which are used by the \textbf{evaluation agent} to assess a generated image and produce both factuality judgments and textual feedback.
    Finally, the feedback is used for \textbf{feedback-driven image improvement}, yielding a more factual revised image.
    For clarity, the figure shows only representative examples rather than the full intermediate outputs.}
    \label{fig:2}
\end{figure*}

\subsection{What is a fact in T2I generation?}
\label{sec:3.1}
Cambridge Dictionary defines a \textit{fact} as ``something that is known to have happened or to exist, especially something for which proof exists.'' 
Building on this notion, we focus on the T2I setting where evaluation must rely on what can be visually checked in an image. 
Accordingly, we define a \textit{fact} as ``information can be objectively verified from visual evidence, possibly together with reliable sources grounded in the prompt.''

This distinction is important because not all plausible or commonly used attributes should be treated as facts. 
For example, for the prompt \textit{``a water molecule,''} fact includes properties such as the presence of two hydrogen atoms and one oxygen atom, as well as their bent molecular structure. 
By contrast, conventional depictions should not be treated as facts: \textit{``oxygen atom is red''} or \textit{``hydrogen atoms are labeled with H''} are not a fact.


Importantly, factual requirements are also prompt-dependent. 
For example, \textit{``three dogs''} only requires general visual properties of dogs, whereas \textit{``three French bulldogs''} additionally implies breed-specific facial and body characteristics. 
Likewise, temporal context can change the factual attributes of the same object. 
For the prompt \textit{``the Statue of Liberty in 1890,''} the statue should appear copper-brown rather than its present-day green patina, since the oxidation process had not yet produced in 1890.

These examples illustrate that facts in T2I generation are subtle and often under-specified, yet humans can still distinguish them reliably. 
One reason is that human visual perception is structured: people first grasp the global structure and then attend to finer details~\cite{schyns1994blobs, bar2006top}.
Motivated by this coarse-to-fine process, we organize facts into three levels.

\subsection{Three-level fact taxonomy}
We hierarchically structure factual information from coarse semantic consistency to fine-grained details:

\noindent\textbf{Level 1: Object identity and scene recognition.}
This level verifies whether the image depicts the core subject and overall scene described by the prompt.

\noindent\textbf{Level 2: Key component verification.}
This level checks whether the image contains the major attributes, parts, or structural properties that support the identity of the subject.

\noindent\textbf{Level 3: Fine-grained detail verification.}
This level evaluates more specific details and atomic factual properties that are necessary for full factual correctness.

In addition to this three-level hierarchy, we categorize each fact into one of nine semantic categories---\textit{existence, counting, relation, shape, size, color, posture, scene}, and \textit{others}---following prior work on structured image evaluation~\cite{hu2023tifa, li2024genai, lim2025evaluating}.

For example, consider the prompt \textit{``the Statue of Liberty in 1890.''}
At Level 1, we verify the main object and the overall scene at a coarse level, \eg, whether the Statue of Liberty (or a visually similar statue) is present.
At Level 2, we verify key identity-supporting attributes, including color (\eg, a copper-colored appearance), posture (\eg, the right arm raised holding a torch), relation (\eg, the right hand holding the torch and the left hand holding the tablet).
At Level 3, we verify fine-grained details, such as whether the crown has seven spikes and whether the tablet bears the inscription \textit{``JULY IV MDCCLXXVI.''}

\subsection{Agent-based factual evaluation pipeline}
As described in \cref{fig:2}, We build an agent-based pipeline that progressively transforms an underspecified prompt into a set of verified, visually grounded factual questions, and then uses them to evaluate and improve generated images.
Our framework consists of several specialized language-model-based agents with complementary roles, followed by a feedback-driven image improvement module.
(1) A fact proposal agent leverages the prior knowledge of a LLM to extract candidate visual facts implied by the prompt.
(2) A reference-guided fact extraction agent then analyzes one or more reference images to collect directly observable visual elements associated with the prompt.
(3) A verification agent consolidates these two sources, removing unsupported or non-essential requirements while adding missing identity-defining facts when necessary, thereby producing a verified factual rubric.
Based on this rubric, (4) a QA agent converts each fact into QA pairs, and (5) a VLM-based evaluation agent uses them to assess generated images and produce both scores and natural-language feedback.
Finally, this feedback is fed into image generation or editing model, yielding a fully training-free evaluation-and-improvement pipeline.
In our implementation, the fact proposal agent, verification agent, and QA generation agent are LLM-based and use GPT-5.4-mini~\cite{openai_gpt54mini_2026}, while the reference-guided fact extraction agent and the evaluation agent are VLM-based and use Qwen3-VL-8B-Instruct~\cite{bai2025qwen3}.

\noindent\textbf{Fact proposal agent.}
The fact proposal agent is an LLM-based agent that constructs a factual proposal from the prompt by leveraging the prior knowledge encoded in the agent.
It preserves facts explicitly stated in the prompt and additionally proposes prompt-grounded, visually verifiable facts that are necessary for faithful depiction.
To ensure structured outputs, the agent organizes facts into our three-level hierarchy and nine semantic categories.

\noindent\textbf{Reference-guided fact extraction agent.}
The initial factual proposal may be incomplete for two reasons.
First, the relevant facts may not be fully covered by the prior knowledge available to the proposal agent. Second, some critical visual properties may not be sufficiently specified by text alone.
To address this, the reference-guided fact extraction agent analyzes a reference image from benchmark and extracts directly visible visual elements that can support factual verification.
It is explicitly constrained to rely only on image evidence.
The extracted elements are conservative and visually grounded, focusing on directly observable properties such as color, shape, layout, and distinctive components.
These cues are particularly useful for prompts involving specific products, historical objects, official designs, or culture-specific concepts, where appearance plays an important role in factual identification.

\noindent\textbf{Verification agent.}
The factual proposal and the reference-derived visual elements are then passed to the verification agent, which constructs the final factual rubric for downstream evaluation.
The verification agent treats the prompt as the ground-truth anchor for what is required, while using the reference-derived elements as supporting visual evidence for how the prompt appears.
It verifies whether each proposed fact is necessary, visually verifiable, and relevant to the prompt, and it may add missing facts when they are semantically implied by the prompt or correspond to identity-defining visual properties.
At the same time, it removes elements that are unnecessary, not visually verifiable, or based only on depiction conventions or incidental visual attributes.
The output of this stage is a verified rubric together with explicit add/drop decisions, making the refinement process interpretable.

\noindent\textbf{QA agent.}
Given the verified factual rubric, the QA agent converts each fact into a question.
Each fact is mapped to exactly one atomic QA pair and the intended answer for a correct image is always \textit{yes}.
To make the evaluation interpretable and robust, the agent is instructed to keep each question short, concrete, and visually answerable, while remaining faithful to the source rubric.
Importantly, the agent is designed to reduce repeated penalization by making each question as independent as possible.
Otherwise, uncertainty about object identity can propagate to many downstream questions. 
For example, if the evaluator is unsure whether an object is an apple, related questions such as \textit{``Is the apple on the table?''} may also be scored incorrectly, even though object existence is already evaluated separately by a question such as \textit{``Is there an apple?''}
We therefore phrase questions to preserve the underlying target while reducing unnecessary dependence on object identity.
For example, when evaluating a ``relation'', we may ask \textit{``Is the main object on the table?''} instead of \textit{``Is the apple on the table?''}.
The resulting QA set preserves the level and category metadata of the factual rubric and stores the corresponding facts for later interpretation and feedback generation.

\begin{figure*}[t]
    \centering
    \includegraphics[width=\textwidth]{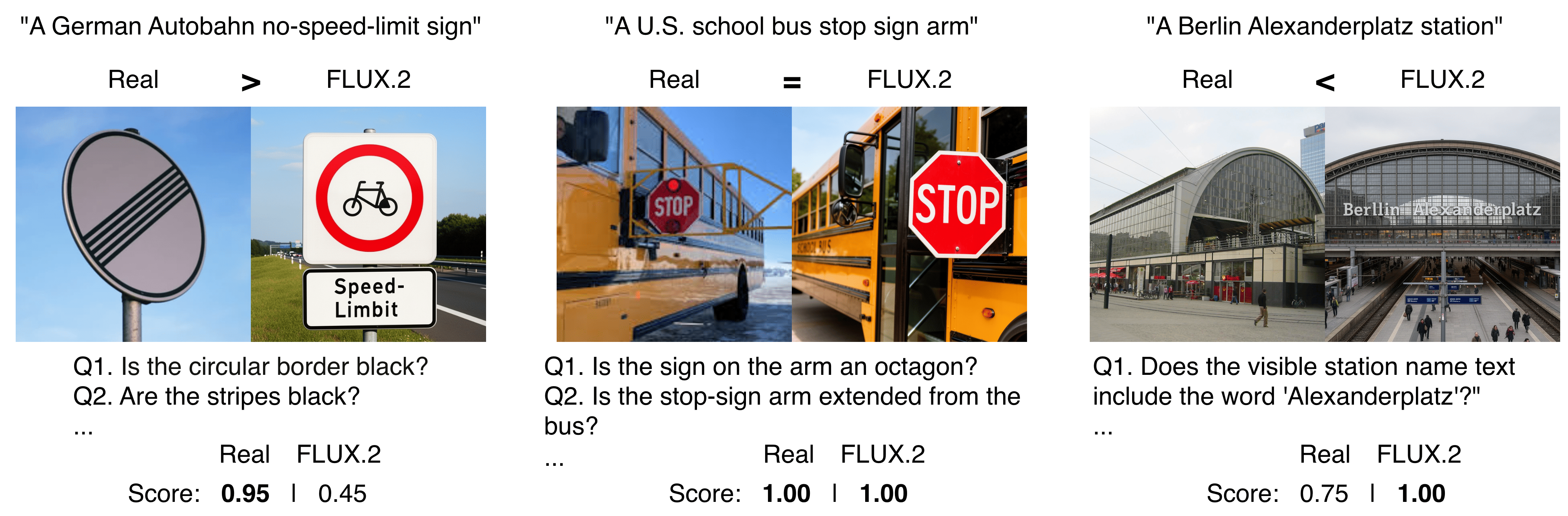}
    \caption{\textbf{Illustration of the Factual A/B test with FAGER score.} For each prompt, we compare the FAGER scores of a real image and a generated image. The left example shows a correctly ranked case where the real image scores higher. The middle example shows a tie, which we count as correct because the generated image may also satisfy the factual requirements of the prompt. The right example shows a failure case where the real image scores lower because the required fact is not visibly verifiable in the real image itself.}
    \label{fig:factual_ab_examples}
\end{figure*}

\noindent\textbf{Evaluation agent.}
\label{sec:3.3}
Given a generated image and its QA set as input, the evaluation agent outputs question-level answers with rationales, an overall factuality score (FAGER score), a decision (among \textit{keep}, \textit{edit}, and \textit{regenerate}), and corresponding textual feedback for improvement.

It performs factual assessment in a question-answering manner.
For each question, the agent must answer using only \textit{yes}, \textit{no}, or \textit{unknown}, together with a one-sentence rationale grounded strictly in visible evidence.
The agent is explicitly constrained to use only the image itself and not any external or prior knowledge. If the relevant evidence is not visually verifiable due to factors such as occlusion, blur, small size, or viewpoint, it must answer \textit{unknown}.
We assign scores of $1$, $0$, and $0.5$ to \textit{yes}, \textit{no}, and \textit{unknown}, respectively, and compute the factuality score as the average over all evaluated questions.

The evaluation follows the same coarse-to-fine hierarchy as our factual taxonomy.
We first evaluate only the Level~1 questions.
If the Level~1 score falls below a \textit{regeneration threshold}, the image is deemed to have failed to capture the prompt's core object or scene, and the agent outputs \textit{regenerate} without evaluating Level~2 or Level~3.
Otherwise, the agent continues to evaluate the remaining levels and aggregates all answers into an overall score.
If the final score exceeds a \textit{keep threshold}, the image is accepted with a \textit{keep} decision; otherwise, it is marked as \textit{edit}.

In this paper, we use fixed thresholds across all prompts and datasets: a regeneration threshold of $20$ and a keep threshold of $95$. These values were chosen as practical operating points: the regeneration threshold conservatively identifies clear semantic failures that are better handled by regeneration, while the keep threshold sets a high bar for accepting near-factual images. We use the same thresholds across all datasets to avoid domain-specific tuning. In principle, however, these thresholds may be adjusted depending on the application domain or desired error tolerance, and threshold calibration is a promising direction for future work.

In addition to scores, the evaluation agent also produces textual feedback.
When the decision is \textit{regenerate}, the agent generates a regeneration constraint to be appended to the original prompt, \eg, \textit{``the Statue of Liberty in an outdoor harbor setting.''}
When the decision is \textit{edit}, it generates an edit instruction specifying only the factual changes needed, \eg, \textit{``change the statue color to copper-brown.''}
No additional feedback is generated for \textit{keep}.

\subsection{Feedback-driven image improvement}


Given the decision and textual feedback from the evaluation agent, the system performs one of three refinement actions: (1) keep, (2) edit, or (3) regenerate. 
As described in \cref{sec:3.3}, regeneration is guided by the generated regeneration prompt, whereas editing is guided by the generated edit instruction. The resulting image may be re-evaluated for further refinement, enabling an iterative improvement process; however, in this paper, we perform only a single refinement round. Because this stage is decoupled from the evaluator, FAGER can be combined with different image generation and editing models.


\section{Experiments}

\begin{table*}[t]
\caption{Pairwise accuracy on the Factual A/B test across datasets. Higher is better. For each prompt, a metric is considered correct if it ranks the factual image at least as favorably as the generated image according to its scoring direction. Numbers in parentheses denote the number of prompt--image pairs. Higher is better. Best results are shown in bold.}
\label{tab:metric_comparison}
\centering
\small
\begin{tabular}{lccccc}
\toprule
\textbf{Metric} 
& \textbf{I-HallA-Science (99)} 
& \textbf{I-HallA-History (99)} 
& \textbf{ABO (50)} 
& \textbf{Culture (30)} 
& \textbf{T2I-FactualBench-SKCM (100)} \\
\midrule
\footnotesize
VQAScore~\cite{lin2024evaluating}     & 0.37     & 0.53     & 0.38    & 0.47     & 0.37          \\
FineGRAIN~\cite{hayes2025finegrain}   & 0.56     & 0.72     & 0.68    & 0.83     & 0.76          \\
FAGER  (Ours)      & \textbf{0.73} & \textbf{0.83} & \textbf{0.82} & \textbf{0.97} & \textbf{0.87} \\
\bottomrule
\end{tabular}
\end{table*}

\begin{table*}[t]
\caption{Factuality performance of T2I models and FAGER-guided refinement across datasets, \textbf{measured by FAGER}. Numbers in parentheses denote the number of prompt--image pairs. Higher is better. Best results are shown in bold; second-best results are underlined.}
\centering
\footnotesize
\begin{tabular}{lccccc}
\toprule
\textbf{Model} 
& \textbf{I-HallA-Science (99)} 
& \textbf{I-HallA-History (99)} 
& \textbf{ABO (50)} 
& \textbf{Culture (30)} 
& \textbf{T2I-FactualBench-SKCM (100)} \\
\midrule
FLUX1-dev~\cite{blackforestlabs2024flux1dev}         & 66.99 & 74.58 & 62.16 & 60.19 & 60.19 \\
Stable Diffusion 3.5 Large~\cite{sd35_huggingface_2024}  & 51.32 & 68.89 & 65.20 & 66.75 & 67.17 \\
FLUX2-dev~\cite{flux-2-2025}      & 63.16 & 76.92 & 85.17 & 83.36 & 82.81 \\
Nano Banana Pro~\cite{nanobanana}    & \textbf{90.83} & \textbf{87.76} & 82.26 & 86.53 & \textbf{88.93} \\
\midrule
FLUX1-dev + FAGER (Qwen-Edit) & \underline{76.36} & 79.44 & \underline{85.74} & \underline{87.97} & \underline{79.09} \\
FLUX1-dev + FAGER (Kontext) & 74.32 & \underline{81.20} & \textbf{88.23} & \textbf{89.40} & 79.92 \\
\bottomrule
\end{tabular}
\label{tab:model_results}
\end{table*}

We evaluate FAGER from three perspectives listed below:
\begin{enumerate}
    \item \textbf{Validate FAGER as a factuality metric:} we propose a factual A/B test, which measures whether a metric can give higher score to factually correct images and distinguish them from nonfactual ones. This directly tests whether the metric captures prompt-grounded factual information. (Sec.~\ref{sec:factual_ab_test}, \cref{sec:validation}).
    \item \textbf{Use FAGER to assess the factuality of existing T2I models:} We test across diverse datasets covering scientific knowledge, historical facts, product appearance, culture-specific concepts, and broader visually grounded facts~\cite{lim2025evaluating, collins2022abo, huang2024t2i}. (Sec.~\ref{sec:generation}).
    \item \textbf{Leverage FAGER to provide actionable feedback:} we show that FAGER can guide image refinement, enabling substantially more factual image generation (Sec.~\ref{sec:generation}).
\end{enumerate}

\paragraph{Datasets}
We conduct experiments on five datasets spanning diverse forms of visually grounded factual knowledge: \textbf{I-HallA-Science} and \textbf{I-HallA-History}~\cite{lim2025evaluating}, \textbf{ABO}~\cite{collins2022abo}, \textbf{Culture}, and \textbf{T2I-FactualBench-SKCM}~\cite{huang2024t2i}.
\begin{itemize}
    \item \textbf{I-HallA} was introduced to evaluate hallucination in T2I generation and consists of prompts, real images, and hallucinated images generated by DALL-E~3~\cite{dalle3}. From this benchmark, we use 99 prompts each from science and history subsets, along with their corresponding real images and hallucinated DALL-E~3 images.
    \item  \textbf{ABO} (Amazon Berkeley Objects) contains catalog product images paired with specific metadata, including attributes such as brand or color information. We randomly sample 50 products and use the corresponding real product images as factual references, while generating comparison images with FLUX.2-dev~\cite{flux-2-2025}.
    \item  \textbf{Culture} is our newly constructed dataset targeting culturally specific concepts with visually distinctive properties. We generate candidate prompts with GPT-5.2~\cite{openai2025gpt52systemcard} and retain 30 prompts for which three annotators unanimously agree on their validity. We then retrieve candidate reference images through Google Image Search and select one factual reference image per prompt only when all three annotators unanimously agree. The corresponding generated images are produced with FLUX.2-dev.
    \item \textbf{T2I-FactualBench-SKCM} is a dataset consisting of single-object prompts and corresponding real images. We randomly sample 100 prompts and corresponding real images from this subset, and generate comparison images with FLUX.2-dev.
\end{itemize}

\subsection{Factual A/B Test}
\label{sec:factual_ab_test}

Evaluating a factuality metric is inherently challenging because there is no ground-truth scalar score for image factuality. Although human evaluation is possible, it is expensive and often requires domain expertise, especially for prompts involving science, history, products, or culture-specific knowledge. 
To validate FAGER as a factuality metric, we propose a \textbf{Factual A/B test}, which examines whether a metric prefers factual reference images over corresponding generated images. This directly tests whether the metric captures facts grounded in the prompt.

Specifically, for each prompt, we construct an image pair consisting of a factual reference image $I_{\text{factual}}$ and a generated image $I_{\text{generated}}$.
Since generated images may also be factually correct, we regard a pair as correctly ranked, as long as the factual reference image is not scored lower than the generated image.
For metrics where higher scores indicate better factuality, we define pairwise correctness as
\begin{equation}
\mathrm{correct} =
\begin{cases}
1, & \text{if } s(I_{\text{factual}}) \ge s(I_{\text{generated}}),\\
0, & \text{otherwise},
\end{cases}
\end{equation}
where $s(\cdot)$ denotes the score assigned by the metric. For metrics where lower scores are better, the inequality is reversed. 
We treat ties as correct because generated images are sometimes factually correct and may satisfy the factual constraints of the prompt.
As illustrated in \cref{fig:factual_ab_examples}, this A/B test includes correctly ranked cases, ties counted as correct, and failure cases where the factual reference (real) image is scored lower.
The final \textbf{pairwise accuracy} is computed as the fraction of correctly ranked pairs over the full dataset.

This protocol offers two advantages. First, it evaluates a minimal requirement of a factuality metric: a factual reference image should not be ranked worse than a generated image for the same prompt. Second, it enables fair comparison across metrics with different scales, since it depends only on relative ranking rather than absolute calibration.

\begin{figure*}[t]
    \centering
    \includegraphics[width=\textwidth]{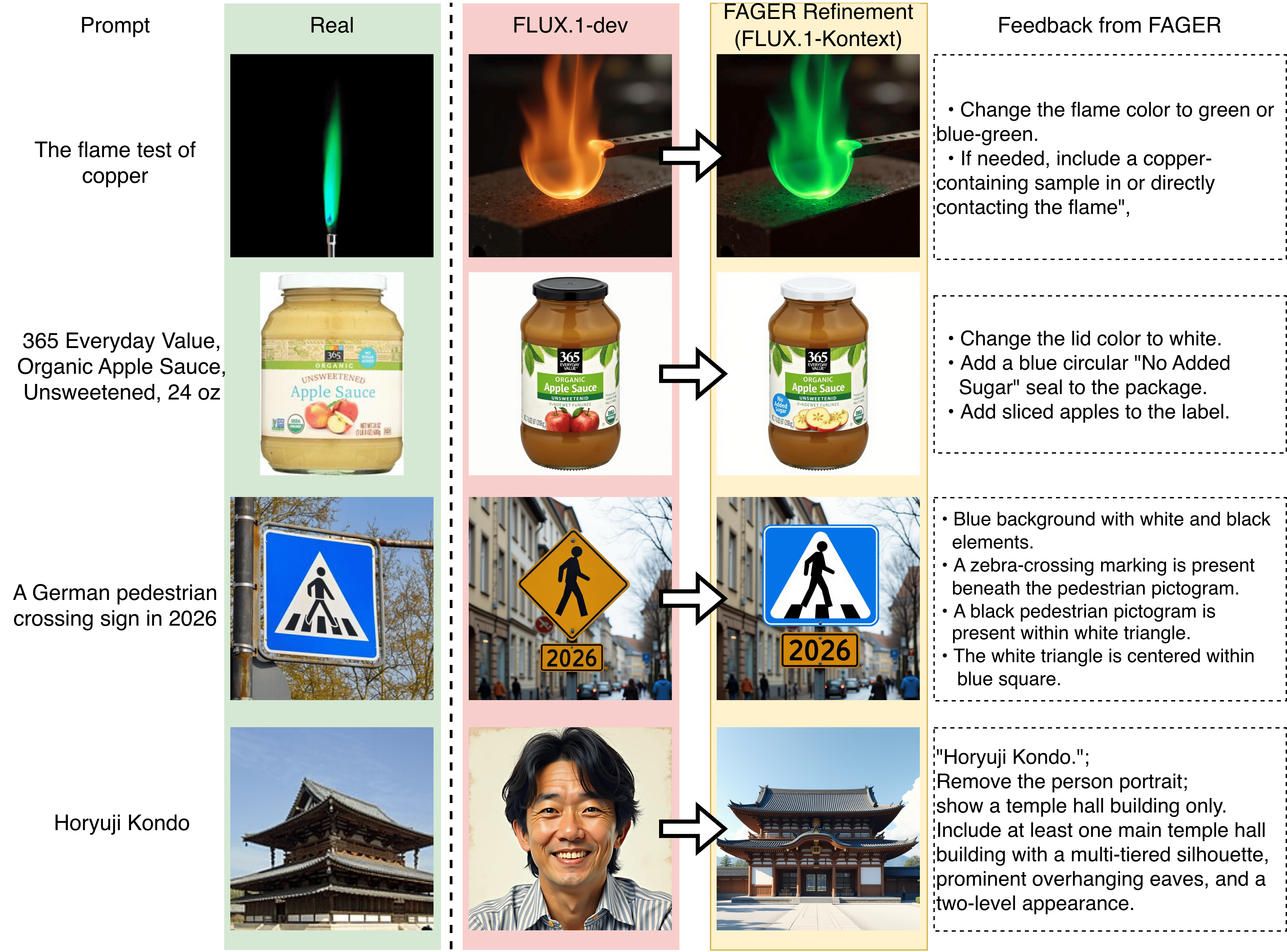}
    \caption{\textbf{Qualitative examples of FAGER-guided image refinement.} The first three rows show \textit{edit} cases, where FAGER identifies localized factual errors and produces targeted edit instructions that improve factual correctness while preserving the overall image content. These examples include correcting the flame color in a copper flame test, revising product-specific packaging details, and fixing the visual structure of a traffic sign. The last row shows a \textit{regenerate} case, where the original image fails to depict the intended object category and FAGER instead provides regeneration feedback to guide a more faithful result.}
    \label{fig:refinement_examples}
\end{figure*}

\subsection{Metric Validation}
\label{sec:validation}

\cref{tab:metric_comparison} reports pairwise accuracy on the Factual A/B test across all datasets. 
For comparison, we evaluate representative prior metrics with different scoring mechanisms. VQAScore~\cite{lin2024evaluating} reformulates the prompt as a single yes/no question, and uses the VQA model's predicted probability of answering ``Yes'' as the 0 score. 
FineGRAIN~\cite{hayes2025finegrain}, one of the most recent VQA-based T2I evaluation metrics, evaluates a broad taxonomy of 27 categories. 
Since lower FineGRAIN scores indicate better factuality, we take it's inverse value when computing pairwise accuracy.

From \cref{tab:metric_comparison}, FAGER consistently outperforms all cases, achieving $0.73$ on I-HallA-Science, $0.83$ on I-HallA-History, $0.82$ on ABO, $0.97$ on Culture, and $0.87$ on T2I-FactualBench-SKCM. 
In contrast, VQAScore performs substantially worse across all datasets, while FineGRAIN is more competitive but remains consistently below FAGER. 
These results indicate that FAGER reliably ranks factual images over corresponding generated images, suggesting that it better captures prompt-grounded factual information.

To complement the Factual A/B test, we conducted a small-scale human evaluation on ABO and Culture, since datasets such as science often require stronger domain-specific knowledge and are less suitable for lightweight human evaluation. For each dataset, we randomly sampled 10 prompt--image pairs and recruited five annotators with undergraduate degrees. Annotators rated the factual correctness of each factual reference image and generated image on a 1--5 scale, with external image search allowed when needed. We converted the averaged ratings into pairwise preferences and measured metric agreement using the same rule as in the Factual A/B test. FAGER agreed with human judgments on $80.0$\% of ABO pairs and $90.0$\% of Culture pairs, compared to $40.0$\% and $80.0$\% for FineGRAIN. Although limited in scale, these results further support that FAGER is better aligned with human factuality judgments.

\subsection{FAGER for refinement of generations}
\label{sec:generation}

Using the same prompts as in \cref{tab:metric_comparison}, we evaluate the factuality of images generated by different T2I models using FAGER. We further assess how much the factuality of FLUX1-dev~\cite{blackforestlabs2024flux1dev} outputs can be improved through FAGER-guided refinement. In all cases, images assigned \textit{regenerate} are regenerated with FLUX1-dev. For images assigned \textit{edit}, we consider two editing backbones, Qwen-Image-Edit-2511 (Qwen-Edit)~\cite{qwen_image_edit_2024} and FLUX.1-Kontext-dev~\cite{blackforestlabs2025kontext}, and report results for both variants.

As shown in \cref{tab:model_results} and \cref{fig:refinement_examples}, FAGER-guided refinement consistently improves the factuality of FLUX1-dev outputs across all datasets. FLUX1-dev + FAGER (Kontext) improves over the original FLUX1-dev from $74.58$ to $81.20$ on I-HallA-History, from $62.16$ to $88.23$ on ABO, and from $60.19$ to $89.40$ on Culture, achieving the best results on ABO and Culture. FLUX1-dev + FAGER (Qwen-Edit) also yields substantial gains, improving I-HallA-Science from $66.99$ to $76.36$ and performing competitively across the remaining datasets. On T2I-FactualBench-SKCM, both refinement variants substantially outperform FLUX1-dev.

\begin{figure}[t]
    \centering
    \includegraphics[width=1.0\linewidth]{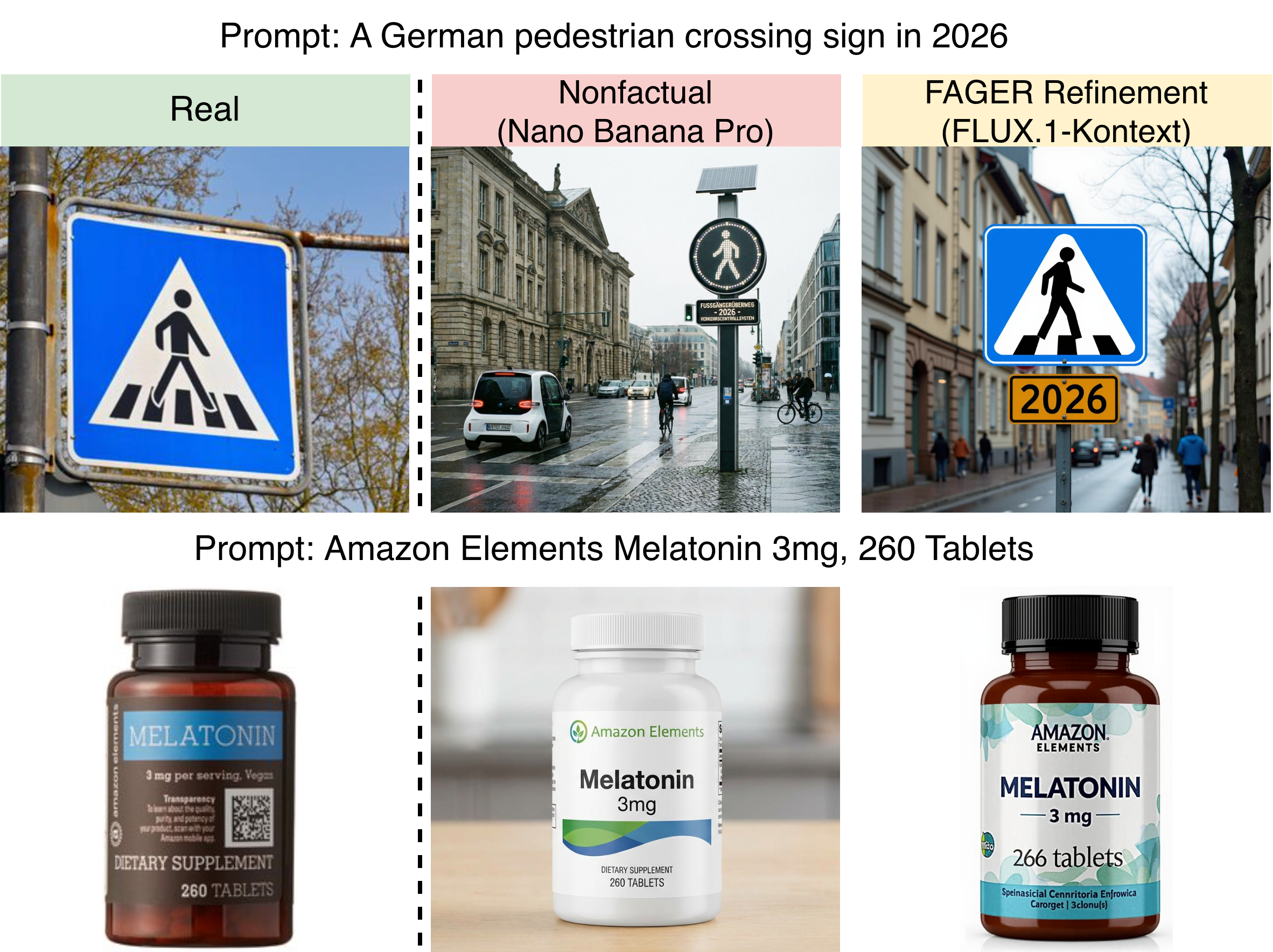}
    \caption{Qualitative examples where FAGER-guided refinement achieves higher factuality than Nano Banana Pro. Prompts are drawn from the ABO and Culture datasets.}
    \label{fig:fig5}
\end{figure}

Notably, Nano Banana Pro~\cite{nanobanana} achieves the best performance on I-HallA-Science, I-HallA-History, and T2I-FactualBench-SKCM. In contrast, FAGER-guided refinement achieves the best results on ABO and Culture, with \cref{fig:fig5} illustrating qualitative cases where it yields more factual outputs than Nano Banana Pro. These results show that FAGER provides effective and actionable feedback, enabling a weaker base model to approach stronger generators on several datasets without any additional training.

Moreover, the strong performance of the Qwen-Edit variant shows that our refinement pipeline is not limited to the FLUX family and can be applied more broadly with different image editing backbones.

\cref{fig:refinement_examples} provides qualitative examples of FAGER-guided refinement. The top three rows show \textit{edit} cases, where FAGER identifies missing or incorrect facts and produces edit instructions that improve factual correctness while preserving the overall image content. 
The bottom row shows a \textit{regenerate} case, where the original generation fails to depict the intended object and FAGER instead provides regeneration feedback to guide a more faithful result.


\begin{table}[t]
\caption{Average CLIPScore and LPIPS across all five datasets. 
Higher is better for CLIPScore, while lower is better for LPIPS.}
\centering
\footnotesize
\begin{tabular}{lcc}
\toprule
\textbf{Method} & \textbf{CLIPScore~\cite{hessel2021clipscore}} $\uparrow$ & \textbf{LPIPS~\cite{zhang2018unreasonable}} $\downarrow$ \\
\midrule
FLUX1-dev & 30.92 & 0.7407 \\
FLUX2-dev & 31.98 & 0.7245 \\
FLUX1-dev + FAGER (Kontext) & 30.78 & 0.7372 \\
\bottomrule
\end{tabular}
\label{tab:clip_lpips_avg}
\end{table}

We additionally report CLIPScore~\cite{hessel2021clipscore} and LPIPS~\cite{zhang2018unreasonable} as auxiliary metrics to assess prompt alignment and perceptual similarity. Although these metrics do not measure factual correctness, they help verify that refinement does not degrade image quality. As shown in \cref{tab:clip_lpips_avg}, FAGER-guided refinement preserves CLIPScore relative to FLUX1-dev and maintains comparable, or slightly improved, LPIPS to real reference images. These results indicate that factuality gains are achieved without compromising alignment or perceptual quality.

\subsection{Limitations and component reliability}
\label{sec:limitations}

FAGER relies on multiple LLM- and VLM-based components, so its reliability can be affected by upstream hallucinations or recognition errors. We reduce these risks by cross-checking prior-based and reference-based information, explicitly removing unsupported or convention-based attributes in the verification stage, and allowing the evaluator to output \textit{unknown} when a fact is not visually verifiable. Moreover, our Factual A/B test provides end-to-end validation of the full pipeline, showing that despite imperfect components, FAGER yields more reliable factuality judgments than prior metrics. Nevertheless, robustness to component-level failures remains an important limitation and a promising direction for future work.

\section{Conclusion}

We introduced \textbf{FAGER}, a training-free framework for factually grounded evaluation and refinement in T2I generation. This work takes a step toward formalizing what should count as a \emph{fact} in T2I generation, an aspect that has remained largely underexplored. We formalize factuality through a three-level taxonomy, capturing object identity, key attributes, and fine-grained details, and use it to build an agent-based pipeline for both evaluation and improvement.

FAGER constructs verified factual rubrics, converts them into interpretable QA sets, and uses them to assess images and generate actionable feedback. We also propose a \textbf{Factual A/B test} for validating factuality metrics, and show across five datasets that FAGER consistently outperforms prior approaches. Beyond evaluation, FAGER improves image generation through targeted editing and regeneration while preserving prompt alignment and perceptual quality.

Overall, our results highlight that factuality in T2I requires going beyond prompt alignment to explicitly verify implicit factual constraints. We hope FAGER contributes toward more reliable evaluation and more factually faithful image generation.
\newpage
{
    \small
    \bibliographystyle{ieeenat_fullname}
    \bibliography{main}
}


\end{document}